\documentclass{Interspeech2024}
\usepackage{times}
\usepackage{latexsym}
\usepackage{graphicx}
\usepackage{booktabs}
\usepackage{multirow}
\usepackage{algorithmic}
\usepackage{algorithm}
\usepackage{xcolor}
\usepackage{enumitem}
\usepackage{amsfonts}
\setlist[itemize]{noitemsep, topsep=0pt}

\usepackage[utf8]{inputenc}
\usepackage[T1]{fontenc}

\interspeechcameraready

\title{Investigating the 'Autoencoder Behavior' in Speech Self-Supervised Models: a focus on HuBERT's Pretraining.}

\name[affiliation={1}]{Valentin}{Vielzeuf}

\address{
  $^1$ Orange, Cesson-S\'evign\'e, France
\email{valentin.vielzeuf@orange.com}}


\begin{document}

\maketitle

\begin{abstract}
Self-supervised learning has shown great success in Speech Recognition. However, it has been observed that finetuning all layers of the learned model leads to lower performance compared to resetting top layers. This phenomenon is attributed to the "autoencoder" behavior: top layers contain information closer to the input and are less suitable for tasks that require linguistic information, such as Speech Recognition.
To better our understanding of this behavior, we propose to study the evolution of high-level information within the model during pretraining. We focus on the HuBERT model, which exhibits a less pronounced "autoencoder" behavior. By experimentally exploring various factors that may have an impact, we aim to improve the training procedure and enhance the top layers of HuBERT for high-level tasks.
Furthermore, our experiments demonstrate that these improvements in the training procedure result in faster convergence and competitive performance on downstream tasks.

\end{abstract}

\section{Introduction}
 \begin{figure*}[ht!]
    \centering
    \begin{minipage}{0.3\linewidth}
        \caption*{\tiny{Wav2vec2.0 - CCA-Word} }
        \includegraphics[width=\textwidth]{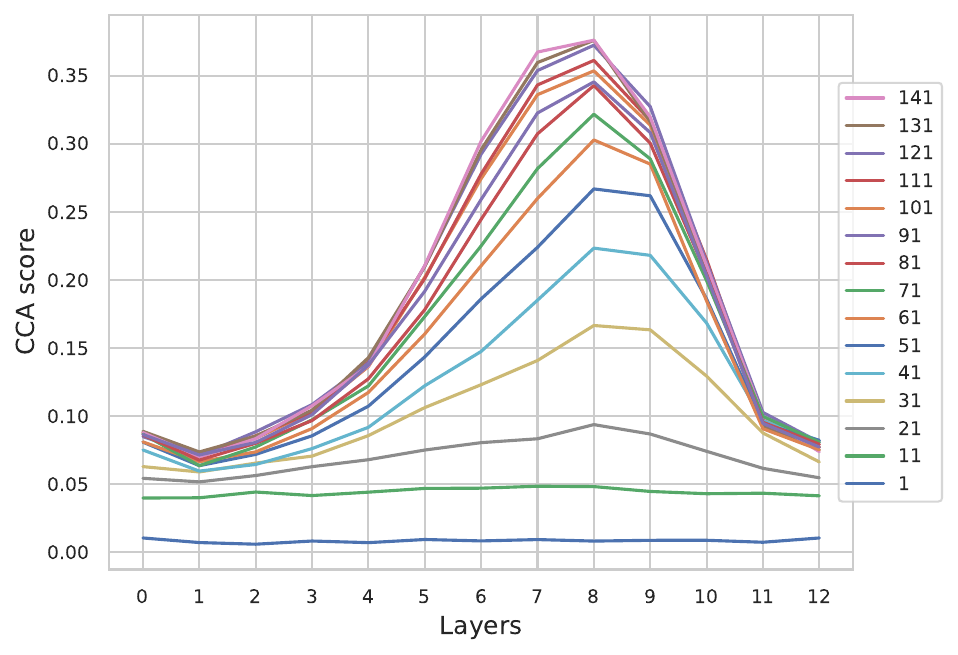}
        
    \end{minipage}
    \centering
    \begin{minipage}{0.3\linewidth}
        \caption*{\tiny{HuBERT First Iteration - CCA-Word} }
        \includegraphics[width=\textwidth]{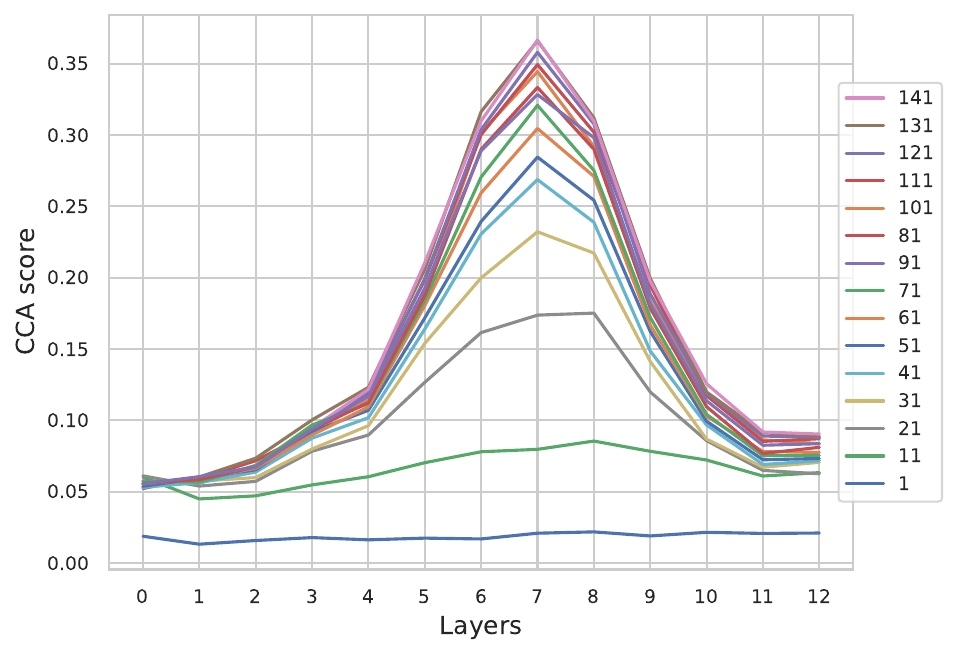} 
        
    \end{minipage}
    \centering
    \begin{minipage}{0.3\linewidth}
        \caption*{\tiny{HuBERT Second Iteration - CCA-Word} }
        \includegraphics[width=\textwidth]{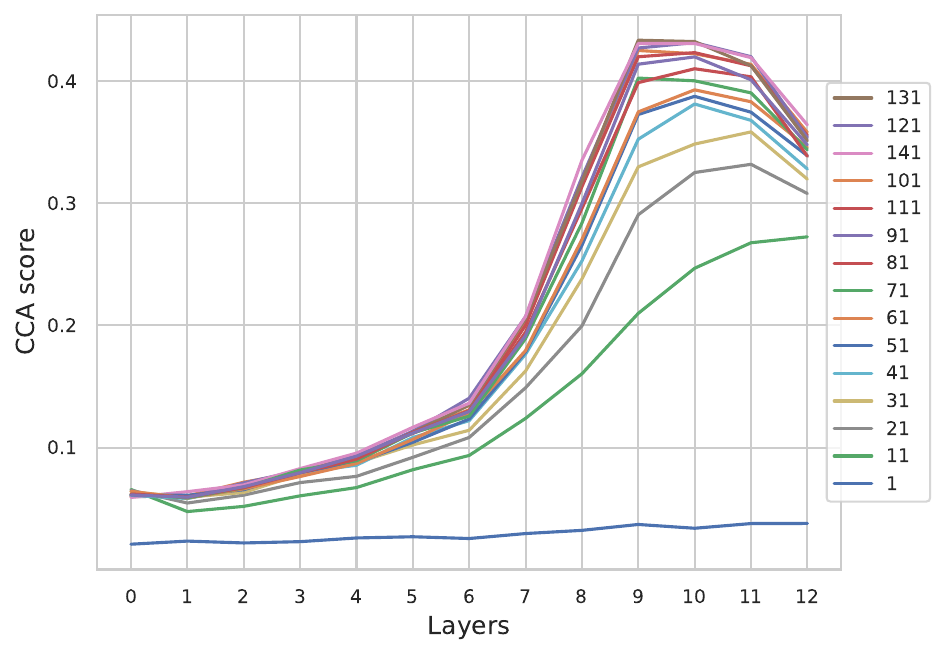}
        
    \end{minipage}
    \centering
\caption{Visualization of the "auto-encoder" behaviour through layerwise analysis of the training of Wav2vec2 and HuBERT. A CCA-score (y-axis) modelling the word-level information contained by each layer (x-axis) is provided for different epochs (1 to 141, legend in the right of the figures).}
\label{fig:preliminary}
\end{figure*}
Supervised machine learning requires labeled training data. This labeling process implies a cost that can become not affordable. The issue is amplified by the increasing sizes of corpora and the growing complexity of tasks.
This might be one of the reasons why self-supervised approaches have been widely adopted in fields such as Natural Language Processing~\cite{kenton2019bert} or Computer Vision~\cite{chen2020simclr}.
In the speech processing field, self-supervised learning approaches improve the performance on several tasks (notably Speech Recognition), with foundation approaches such as the wav2vec serie~\cite{schneider2019wav2vec,baevski2019vq,baevski2020wav2vec}.

However, using proxy tasks to replace the need for supervision also has drawbacks. There is no guarantee that the learned models produce representations containing useful information for solving downstream tasks. 
For example, Wav2vec2 seems to lose high-level information in its top layers~\cite{pasad2021layer}. Thus, when finetuning for Speech Recognition, the parameters of these layers are often reset to random values. Since the pretraining cost of such models is significant, discarding a portion of the trained parameters may be seen as a waste of energy.
More recent studies~\cite{pasad2023comparative} suggest that the "autoencoder" behaviour (having top layers bearing input-level information) is less pronounced in certain categories of self-supervised approaches compared to others. However, the majority of approaches exhibits this phenomenon to varying degrees (see Figure~\ref{fig:preliminary}).

This paper aims to investigate the possible causes of this behavior in the specific case of the HuBERT model~\cite{hsu2021hubert} and explore potential strategies to alleviate this phenomenon. These strategies may help to better exploit the capacity of the implemented architectures.
There are three main contributions: \textbf{(a)} a detailed analysis of the factors that impact the representation of the top layers, \textbf{(b)} a strategy that focuses on the HuBERT iteration parameters to preserve high-level information in the upper layers, which slightly improves performance on downstream tasks, and \textbf{(c)} a faster convergence rate for equivalent downstream performance, which results in a two times shorter pretraining time.

\section{Related work}

\textbf{Improving self-supervised discrete units}
Different lines of work propose to implicitly improve the supervision of the HuBERT model by building richer discrete units than the original ones.
For instance, \cite{nguyen2022word} shows that providing word boundaries information to the model is helpful to improve on downstream semantic tasks, while other work focus on the granularity of the clusters~\cite{elkahky2023coarser,kamper2022word}, or even to learn longer-range labels with tokenizaton methods~\cite{ren2022speech}.
Our paper stays in the scope of the original HuBERT, with discrete clusters computed on windows close to phoneme-level. We  focus on the iteration process and investigate if we can reach high-level information (i.e. related to semantic contents) in the top layers of the model. Yet, the results obtained in this paper are complementary with these lines of work, as a better cluster strategy may help to provide better supervision to the top layers.

\noindent \textbf{Teacher Student approaches}
HuBERT may be seen as a discrete self-distillation approach, with at each iteration a novel teacher extracted from the previous iteration.
Lines of work exploiting a teacher as a stochastic weight average of the student (such as Data2vec~\cite{baevski2022data2vec} or SPIRAL~\cite{huang2021spiral}) are therefore already achieving very numerous "iterations". But by focusing on HuBERT, we can have an easier and discrete control on the factors regulating the interaction between the teacher and the student. 

\noindent \textbf{Layerwise analysis} 
The analysis of the contents of different layers in self-supervised models has been extensively explored in the literature~\cite{niu2022does}. For example, the SUPERB challenge employs probing through finetuning on downstream tasks~\cite{yang2021superb}.
\cite{pasad2021layer,pasad2023comparative,pasad2023self} propose CCA-based similarity measures and the ZeroSpeech challenge~\cite{dunbar2022self} develops metrics for acoustic and word units discovery.
However, these metrics mainly focus on analysing representations extracted by fully pretrained models. In this paper, we consider these methods as promising tools from different areas that can provide complementary insights. Additionally, we propose to study the dynamics of these metrics during the pretraining phase to better our understanding of, for instance, the impact of various self-supervised methods.

\section{Methods}

\textbf{HuBERT}~\cite{hsu2021hubert}, which stands for Hidden unit BERT, aims at performing BERT-like training on speech contents. There is therefore a need to generate discrete labels. Thus an offline clustering step on top of previously extracted features is performed. The general process is summarized in  Algorithm \ref{algo1}.
The HuBERT method consists of three main steps: (a) feature extraction for each audio sample, (b) clustering of the extracted features using K-means and (c) pretraining using the cluster labels from the previous step. These three steps together form one \textit{iteration}.
In the original approach, the base model performs two iterations, with a larger number of minibatch steps in the second \textit{iteration}.
During the first \textit{iteration}, MFCC features are used for clustering. In subsequent \textit{iterations}, clustering features are embeddings extracted from a carefully selected layer of the model trained in the previous iteration.

The pretraining step follows the principle of masked prediction representation learning. 
As described in~\cite{hsu2021hubert}, let $X$ represent a sequence of $T$ audio samples $x_t$ and let $\tilde{X}$ represent the masked sequence, with $M$ denoting the set of indices where values in the sequence are masked.
The masked prediction model $f$ takes $\tilde{X}$ as input and predicts a distribution over the targets $Z$, obtained through the clustering step. This distribution is denoted as $p_f(z_t | \tilde{X},t)$, where $z_t$ represents the cluster label of $x_t$. The cross-entropy loss can be expressed as:
\begin{equation} 
Loss(f;X,M,Z) = \sum_{t \in M} \log p_f(z_t | \tilde{X},t)
\end{equation}
It can be noted that in the original HuBERT approach, the loss function was a weighted sum of the cross-entropy on masked and unmasked parts. However, no significant benefit was reported from the inclusion of unmasked parts in the loss function.

\begin{algorithm}
\small
\caption{HuBERT training procedure. \\ \textit{Note that functions are detailed in the text.}}
\label{algo1}
\begin{algorithmic}
\STATE \textbf{Let} $N \geq 1$
\STATE $iteration \gets 0$
\STATE $teacher \gets \textit{MFCC}$

\WHILE{$iteration \leq  $N} 
    \STATE $features \gets \color{gray}{\textbf{extractFeatures}}($teacher)
    
    \STATE $labels \gets \color{olive}{\textbf{clustering}}($features)
    
    \STATE $teacher \gets \color{brown}{\textbf{pretraining}}($labels)
\ENDWHILE
\end{algorithmic}
\end{algorithm}

\noindent \textbf{How do we evaluate the quality of the extracted representations?}
The SUPERB~\cite{yang2021superb} challenge provides a framework for assessing the quality of speech representations extracted from diverse self-supervised models. This evaluation is done by probing the representations on different downstream tasks. The probing consists in finetuning the model or using a simple classifier on top of the frozen self-supervised encoder for a specific task, such as Speech Recognition, Speaker Identification, or Intent Classification. This evaluation helps to measure if the extracted representations contain task-specific information. By applying this method to all layers of the encoder (layerwise), we can identify the specific layers where the information relevant to each task is located within the model. This layerwise analysis provides some insights on the distribution and utilization of task-related information in the model.

However, it is important to note that probing has computational costs and the conclusions drawn from it can be influenced by the choice of the probing classifier~\cite{zaiem2023speech}. To complement our analysis methodology, we propose to use representation similarity analysis, following the approach of \cite{pasad2021layer,dunbar2022self}. 
In line with the methodology presented in~\cite{pasad2023comparative}, we employ projection-weighted CCA~\cite{morcos2018insights} to measure the similarity between the representations extracted by the model layers and specific reference representations. These reference representations can include one-hot word labels (CCA-word), more complex embeddings such as GloVe (CCA-GloVe)~\cite{pennington2014glove}, or Acoustically Grounded Word Embeddings (CCA-AGWE)~\cite{settle2019acoustically}. By utilizing representation similarity analysis, we retrieve additional information on the relationship between the learned representations and the reference representations, providing a more comprehensive understanding of the knowledge contained within the model.

\noindent \textbf{The "auto-encoder" phenomenon}
As discussed in the introduction and demonstrated in \cite{pasad2023comparative}, several models, including Wav2vec2~\cite{baevski2020wav2vec}, exhibit a peculiar behavior known as the "auto-encoder" phenomenon. This phenomenon refers to the tendency of the last layers of the encoder to retain information that is closer to the input data, compared to the middle layers. Consequently, when utilizing Wav2vec2 for downstream tasks like Speech Recognition, it is common practice within the community to reset the last layers of the model~\cite{pasad2021layer}. While this behavior is also observed in other models, its intensity is relatively less pronounced.
To illustrate this phenomenon, we conducted a preliminary layerwise analysis (see Figure~\ref{fig:preliminary}), comparing the pretraining behaviours of Wav2vec2, and of HuBERT first and second \textit{iterations}. In line with Wav2vec2, the first iteration of HuBERT exhibits a similar "auto-encoder" trend, indicating that the lower layers retain more input-related information. However, in the second \textit{iteration} of HuBERT, we observe a notable shift, suggesting the presence of higher-level information in the upper layers. This observation leads us to believe that the issue is not solely attributed to the training criterion, as discussed in~\cite{pasad2023comparative}, but also to the dynamics of supervision during the training process. Consequently, our proposed method aims to investigate potential "\textit{iteration}" strategies that can mitigate this "auto-encoder" behavior and potentially yield improvements for downstream tasks.

\noindent \textbf{How could we "improve" the content of top layers?}
Based on our observation that the second \textit{iteration} of the Original HuBERT appears to mitigate the "auto-encoder" behavior, we have chosen to focus on a simple hypothesis: increasing the number of iterations (while keeping the total minibatch steps constant) may facilitate the acquisition of more "high-level" supervision.
However, increasing the number of iterations also introduces the opportunity to experiment with various parameters, such as the number of minibatch steps within each iteration, the number of clusters, and the height(s) of the supervision layer(s). This might result in a relatively large exploration space. 

Considering the computational cost associated with training HuBERT, it is not feasible to explore the entire parameter space. Therefore, we have decided to set the number of \textit{iterations} arbitrarily to N=10~\footnote{This number has been chosen based on the heuristic that we want an increased number of \textit{iterations}, but with enough minibatch steps for the model to converge.} and focus on three main strategies that primarily vary the number of minibatch steps per iteration.
\textbf{The first strategy}, denoted as \textit{Uniform}, serves as a naive baseline approach where an equal number of minibatch steps is allocated to each \textit{iteration}.
\textbf{The second strategy}, referred to as \textit{Progressive}, is based on the assumption that the proxy task in the initial iterations is relatively simpler compared to the later iterations. This strategy involves: (a) Performing N=10 \textit{iterations}, (b) Linearly increasing the number of minibatch steps within each iteration to make the last iteration the longest, (c) Gradually increasing the height of the layer used for extracting embeddings for clustering (while keeping the first one as MFCC and the second one as the sixth layer, as described in the Original paper), and (d) Maintaining a constant number of clusters.
\textbf{The third strategy}, known as \textit{Progressive+Cluster}, explores the potential benefits of progressively increasing the number of clusters across \textit{iterations} (at a higher computational cost), similar to the approach taken in the Original paper~\cite{hsu2021hubert}. This strategy follows a similar framework as the \textit{Progressive} strategy, but additionally linearly increases the number of clusters from 100 to 512.
By implementing these three strategies, we aim to investigate different approaches to improve the content of the top layers in HuBERT.

\section{Results}
\begin{figure}[t!]
    \centering
    \includegraphics[width=0.92\linewidth]{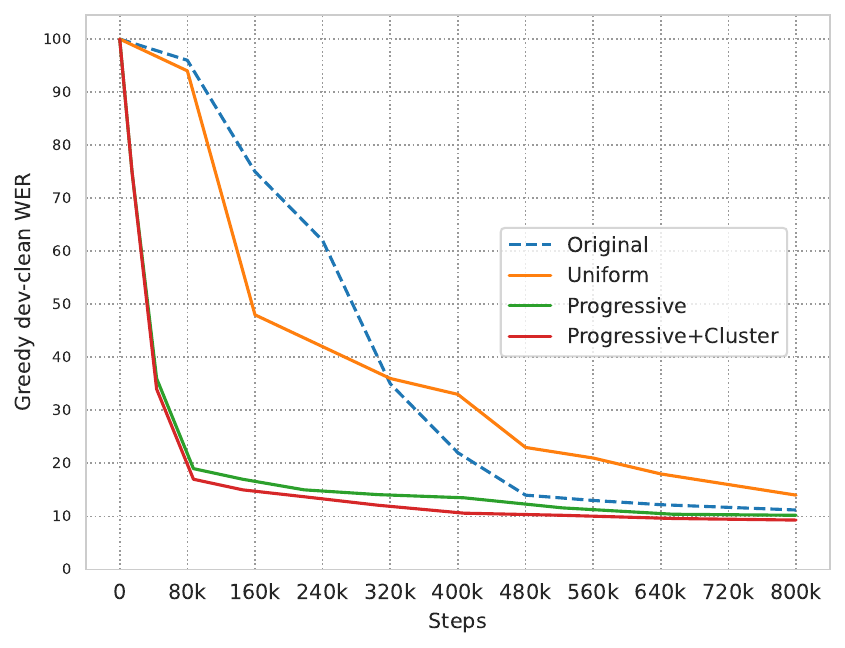}
    \caption{Greedy word error rate evolution (finetuning is performed on LibriLight-10h every 80k pretraining steps) on LibriSpeech Dev-Clean for different pretraining procedures. Comparison with the official HuBERT is done at 650,000 minibatch steps.}
    \label{fig:wer}
\end{figure}

\noindent \textbf{Implementation details}
We pretrain the HuBERT Base model, that is denoted as \textit{Original} in this paper. We base our implementation on \href{https://github.com/pytorch/audio/tree/main/examples/hubert}{torchaudio} and use the hyperparameters proposed in the Original article~\cite{hsu2021hubert}, except for minor changes detailed below.
In order to compute the discrete units clusters on the whole dataset without using a large CPU RAM, we modify the clustering step by implementing real minibatch-kmeans (with partial fitting for each minibatch). Note that the clustering time can be neglected compared to the whole pretraining time.
The pretraining step uses only four 3090 GPUs and therefore accumulate gradients to keep the same virtual batch size. Other hyperparameters are kept unchanged. Finetuning on the LibriLight subsets~\cite{kahn2020libri} and \href{https://github.com/ankitapasad/layerwise-analysis}{layerwise evaluation} are performed following the official repositories.
For the \textit{Uniform} and \textit{Progressive} strategies, we use 100 clusters for all iterations, while for \textit{Progressive+Cluster}, we linearly increase the number of clusters from 100 at the first iteration to 500 at the last. 

\noindent \textbf{ASR probing}
The key finding is that both the \textit{Progressive} and \textit{Progressive+Cluster} strategies yield slightly better WER results compared to the \textit{Original} approach. Additionally, these two methods demonstrate faster convergence. It is worth noting that, for instance, the \textit{Progressive+Cluster} strategy achieves the final WER obtained by the \textit{Original} method, but with a significantly reduced pretraining time of 847 GPU hours compared to the 1900 GPU hours required by the \textit{Original} approach.
As an ablation study, we also observe that the \textit{Uniform} strategy has the opposite effect, slowing down the convergence of the model. This is likely due to allocating excessive time to the initial iterations and not enough to the later ones.
Overall, these findings highlight the effectiveness of the \textit{Progressive} and \textit{Progressive+Cluster} strategies in improving the WER and accelerating the pretraining process compared to the \textit{Original} method.
\begin{table}[t]
\centering
\begin{tabular}{lcccc}
\hline
\multicolumn{1}{c}{\textbf{\textbf{\textbf{Method}}}} & \multicolumn{2}{c}{\textbf{dev}}                & \multicolumn{2}{c}{\textbf{test}}                 \\ \hline
\multicolumn{1}{c}{}                                  & \multicolumn{1}{c|}{clean} & other              & \multicolumn{1}{c|}{clean} & other                \\ \hline
\multicolumn{5}{c}{1 hour Libri-light subset}                                                                                                               \\ \hline
HuBERT~\cite{hsu2021hubert}                           & 5.6                        & 10.9               & 6.1                        & 11.3                 \\[1pt]
Original                                              & 5.9                        & 11.6               & 6.0                        & 11.6                 \\[-3pt]
                                                      & \tiny{[5.32-6.47]}         & \tiny{[10.74-12.32]}  & \tiny{[5.25-6.58]}         & \tiny{[10.64-12.89]} \\[3pt]
Uniform                                               & 7.8                        & 14.3               & 8.4                        & 14.8                 \\[1pt]
Progressive                                           & 5.8                        & 11.1               & 5.9                        & 11.2                 \\[1pt]
Progressive+Cluster                                   & 5.4                        & 10.7               & 6.0                        & 11.4                 \\[1pt] \hline
\multicolumn{5}{c}{10 hours Libri-light subset}                                                                                                             \\[1pt] \hline
HuBERT~\cite{hsu2021hubert}                           & 3.9                        & 9.0                & 4.3                        & 9.4                  \\[1pt]
Original                                              & 4.3                        & 9.5                & 4.4                        & 9.6                  \\[-3pt]
                                                      & \tiny{[3.93-4.52]}         & \tiny{[8.89-9.94]} & \tiny{[4.05-4.77]}         & \tiny{[9.08-10.12]}  \\[3pt]
Uniform                                               & 5.2                        & 10.4               & 5.5                        & 10.7                 \\[1pt]
Progressive                                           & 4.1                        & 9.1                & 4.1                        & 9.2                  \\[1pt]
Progressive+Cluster                                   & 4.0                        & 8.9                & 4.2                        & 9.3                  \\[1pt] \hline
\end{tabular}
\caption{Results when finetuning on LibriLight-1h and LibriLight-10h. Note that our re-implementation of \cite{hsu2021hubert} is reported as Original.}
\label{table:wer}
\end{table}
We have provided a detailed summary of the results in Table~\ref{table:wer}, following the official finetuning procedure with LM decoding. The conclusions drawn from the table align with our previous findings, with one notable difference: increasing the cluster size may have a detrimental effect on generalization.

\begin{figure}
    \centering
    \begin{minipage}{0.7\linewidth}
        \caption*{CCA-Word}
        \includegraphics[width=\linewidth]{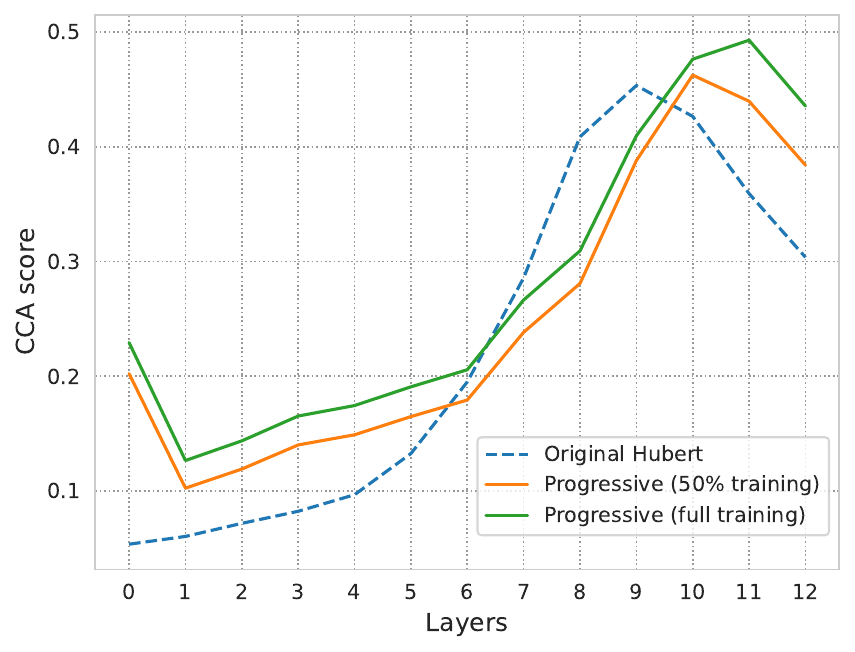}
    \end{minipage}
    \centering
    \begin{minipage}{0.7\linewidth}
        \caption*{CCA-AGWE}
        \includegraphics[width=\textwidth]{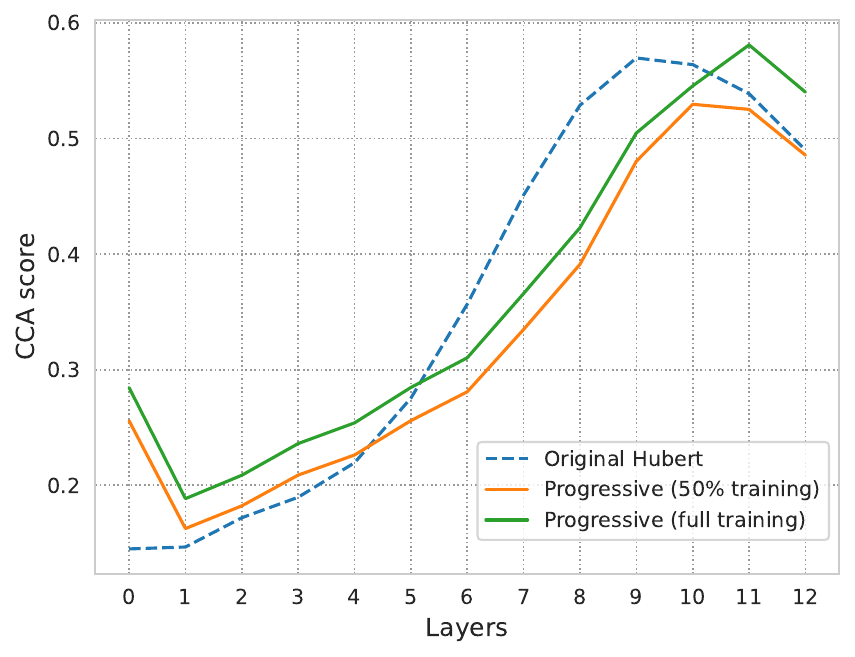}
    \end{minipage}
    \centering
    \begin{minipage}{0.7\linewidth}
        \caption*{CCA-GloVe}
        \includegraphics[width=\textwidth]{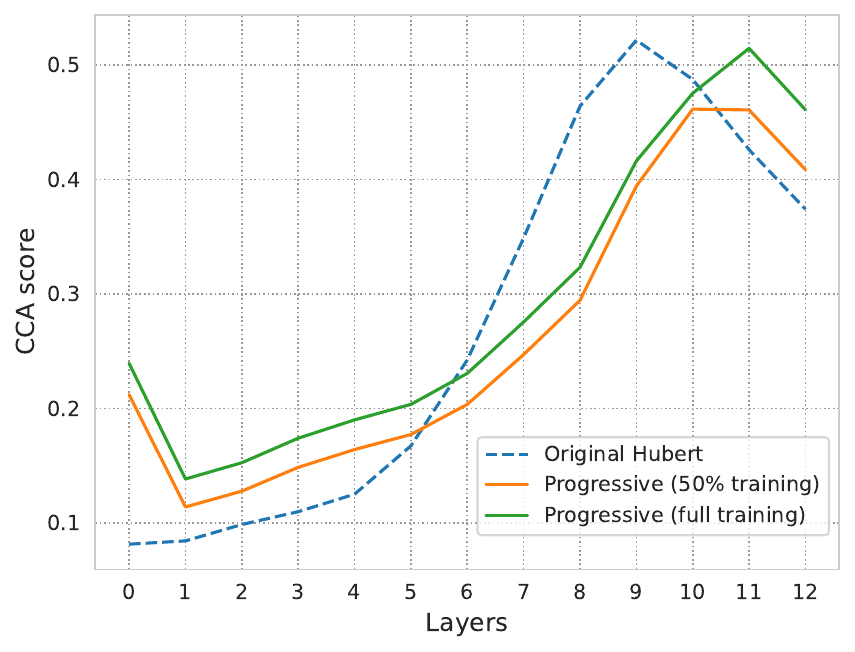}
    \end{minipage}
    
    \centering
    \begin{minipage}{0.7\linewidth}
    \caption*{Intent Classification}
    \includegraphics[width=\linewidth]{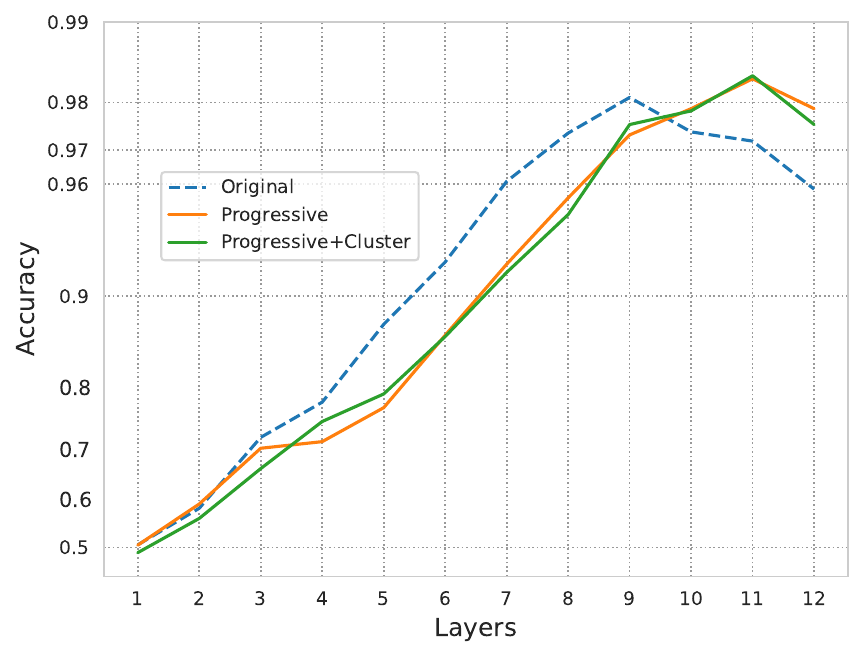}
    \end{minipage}
    
    \caption{First three figures are Layerwise Word, AGWE and GloVe CCA similarites for \textit{Original} and \textit{Progressive} approaches. The last one is the results of probing on Intent Classification (on Fluent Speech Commands dataset~\cite{lugosch2019speech}).}
    \label{fig:layerwise}
\end{figure}
\noindent \textbf{Layerwise analysis}
To assess whether the proposed method has successfully limited the "autoencoder" behavior, we conducted a layerwise analysis of both the \textit{Original} approach and the \textit{Progressive} (at 50\% of the pretraining steps and with full pretraining) approach.
In Figure~\ref{fig:layerwise}, we present the results of this analysis. It is observed that the \textit{Progressive} approach exhibits a higher "best" layer compared to the \textit{Original} approach. For instance, the best layer for the \textit{Progressive} approach is the $11^{th}$ layer, while the best layer for the \textit{Original} approach is the $9^{th}$ layer. However, it is important to note that this observation alone does not provide conclusive evidence that this behavior improves the quality of the obtained representation.
We propose to further investigate the impact of limiting the "autoencdoer" behaviour on the quality of the representation by conducting probing experiments on Intent Classification (using the Fluent Speech Command dataset~\cite{lugosch2019speech}. This task should require representations to contain semantic information. 
Following the protocol outlined in~\cite{yang2021superb}, we evaluated the performance of all layers and report the results in Figure~\ref{fig:layerwise}. We observed a similar trend in terms of the accuracy of the best layer. Furthermore, the best layer achieved by the Progressive approach exhibited the highest accuracy. This suggests that even for tasks that are closer to semantic understanding, there might be a benefit in limiting the "autoencoder" behavior.

\noindent \textbf{Limitations} We choose to limit our study to a specific model, HuBERT base, because of computational constraints. Yet, to further validate the presented conclusions, scaling the model and evaluating other architectures would be needed. Moreover, the study is focusing on LibriSpeech, which contains read speech. More spontaneous speech and more difficult domains might lead to different conclusions.
Finally, the authors did their best to reproduce the official results of HuBERT, but, even if the results are comparable, a performance gap remains. Fairer comparisons would need further investigations\footnote{The gap between our implementation and the official results may be due to the usage of less GPUs during pretraining. Indeed, even with gradient accumulation, the virtual batch size is not strictly equivalent, because modules such as Batch Normalization aggregate statistics during the forward, without taking accumulation into account.} on the causes of these gap.


\section{Conclusions}
We proposed in this paper to investigate on the "autoencoder" phenomenon, observed in self-supervised models such as Wav2vec2. It consists in the fact that top layers representation of the model does not contain the highest-level information and are often closer to the low layers of the model.
We have found that this phenomenon heavily rely on the generated supervision. Moreover, for the specific case of HuBERT, we have shown that it is possible to alleviate this behaviour by finding better pretraining strategies than the original one.
By progressively increasing the size of each HuBERT \textit{iterations} and the height of the supervising layer, we observed a faster convergence (2 times faster than the original model) and even slightly better performance.

\bibliographystyle{IEEEtran}
\bibliography{mybib}

\begin{thebibliography}{10}
\providecommand{\url}[1]{#1}
\csname url@samestyle\endcsname
\providecommand{\newblock}{\relax}
\providecommand{\bibinfo}[2]{#2}
\providecommand{\BIBentrySTDinterwordspacing}{\spaceskip=0pt\relax}
\providecommand{\BIBentryALTinterwordstretchfactor}{4}
\providecommand{\BIBentryALTinterwordspacing}{\spaceskip=\fontdimen2\font plus
\BIBentryALTinterwordstretchfactor\fontdimen3\font minus
  \fontdimen4\font\relax}
\providecommand{\BIBforeignlanguage}[2]{{%
\expandafter\ifx\csname l@#1\endcsname\relax
\typeout{** WARNING: IEEEtran.bst: No hyphenation pattern has been}%
\typeout{** loaded for the language `#1'. Using the pattern for}%
\typeout{** the default language instead.}%
\else
\language=\csname l@#1\endcsname
\fi
#2}}
\providecommand{\BIBdecl}{\relax}
\BIBdecl

\bibitem{kenton2019bert}
J.~D. M.-W.~C. Kenton and L.~K. Toutanova, ``Bert: Pre-training of deep
  bidirectional transformers for language understanding,'' in \emph{NAACL-HLT},
  2019.

\bibitem{chen2020simclr}
T.~Chen, S.~Kornblith, M.~Norouzi, and G.~Hinton, ``Simclr: A simple framework
  for contrastive learning of visual representations,'' in \emph{ICLR}, 2020.

\bibitem{schneider2019wav2vec}
S.~Schneider, A.~Baevski, R.~Collobert, and M.~Auli, ``wav2vec: Unsupervised
  pre-training for speech recognition,'' in \emph{Interspeech}, 2019.

\bibitem{baevski2019vq}
A.~Baevski, S.~Schneider, and M.~Auli, ``vq-wav2vec: Self-supervised learning
  of discrete speech representations,'' in \emph{ICLR}, 2019.

\bibitem{baevski2020wav2vec}
A.~Baevski, Y.~Zhou, A.~Mohamed, and M.~Auli, ``wav2vec 2.0: A framework for
  self-supervised learning of speech representations,'' \emph{NeurIPS}, 2020.

\bibitem{pasad2021layer}
A.~Pasad, J.-C. Chou, and K.~Livescu, ``Layer-wise analysis of a
  self-supervised speech representation model,'' in \emph{ASRU}, 2021.

\bibitem{pasad2023comparative}
A.~Pasad, B.~Shi, and K.~Livescu, ``Comparative layer-wise analysis of
  self-supervised speech models,'' in \emph{ICASSP}, 2023.

\bibitem{hsu2021hubert}
W.-N. Hsu, B.~Bolte, Y.-H.~H. Tsai, K.~Lakhotia, R.~Salakhutdinov, and
  A.~Mohamed, ``Hubert: Self-supervised speech representation learning by
  masked prediction of hidden units,'' \emph{TASLP}, 2021.

\bibitem{nguyen2022word}
T.~A. Nguyen, M.~De~Seyssel, R.~Algayres, P.~Roz{\'e}, E.~Dunbar, and
  E.~Dupoux, ``Are word boundaries useful for unsupervised language learning?''
  \emph{arXiv preprint arXiv:2210.02956}, 2022.

\bibitem{elkahky2023coarser}
A.~Elkahky, W.-N. Hsu, P.~Tomasello, T.-A. Nguyen, R.~Algayres, Y.~Adi,
  J.~Copet, E.~Dupoux, and A.~Mohamed, ``Do coarser units benefit cluster
  prediction-based speech pre-training?'' in \emph{ICASSP}, 2023.

\bibitem{kamper2022word}
H.~Kamper, ``Word segmentation on discovered phone units with dynamic
  programming and self-supervised scoring,'' \emph{TASLP}, 2022.

\bibitem{ren2022speech}
S.~Ren, S.~Liu, Y.~Wu, L.~Zhou, and F.~Wei, ``Speech pre-training with acoustic
  piece,'' in \emph{InterSpeech}, 2022.

\bibitem{baevski2022data2vec}
A.~Baevski, W.-N. Hsu, Q.~Xu, A.~Babu, J.~Gu, and M.~Auli, ``Data2vec: A
  general framework for self-supervised learning in speech, vision and
  language,'' in \emph{ICML}, 2022.

\bibitem{huang2021spiral}
W.~Huang, Z.~Zhang, Y.~T. Yeung, X.~Jiang, and Q.~Liu, ``Spiral:
  Self-supervised perturbation-invariant representation learning for speech
  pre-training,'' in \emph{ICLR}, 2021.

\bibitem{niu2022does}
J.~Niu, W.~Lu, and G.~Penn, ``Does bert rediscover a classical nlp pipeline?''
  in \emph{International Conference on Computational Linguistics}, 2022.

\bibitem{yang2021superb}
S.-w. Yang, P.-H. Chi, Y.-S. Chuang, C.-I.~J. Lai, K.~Lakhotia, Y.~Y. Lin,
  A.~T. Liu, J.~Shi, X.~Chang, G.-T. Lin \emph{et~al.}, ``Superb: Speech
  processing universal performance benchmark,'' in \emph{InterSpeech}, 2021.

\bibitem{pasad2023self}
A.~Pasad, C.-M. Chien, S.~Settle, and K.~Livescu, ``What do self-supervised
  speech models know about words?'' \emph{arXiv preprint arXiv:2307.00162},
  2023.

\bibitem{dunbar2022self}
E.~Dunbar, N.~Hamilakis, and E.~Dupoux, ``Self-supervised language learning
  from raw audio: Lessons from the zero resource speech challenge,'' \emph{IEEE
  Journal of Selected Topics in Signal Processing}, 2022.

\bibitem{zaiem2023speech}
S.~Zaiem, Y.~Kemiche, T.~Parcollet, S.~Essid, and M.~Ravanelli, ``Speech
  self-supervised representations benchmarking: a case for larger probing
  heads,'' \emph{arXiv preprint arXiv:2308.14456}, 2023.

\bibitem{morcos2018insights}
A.~Morcos, M.~Raghu, and S.~Bengio, ``Insights on representational similarity
  in neural networks with canonical correlation,'' \emph{NeurIPS}, 2018.

\bibitem{pennington2014glove}
J.~Pennington, R.~Socher, and C.~D. Manning, ``Glove: Global vectors for word
  representation,'' in \emph{EMNLP}, 2014.

\bibitem{settle2019acoustically}
S.~Settle, K.~Audhkhasi, K.~Livescu, and M.~Picheny, ``Acoustically grounded
  word embeddings for improved acoustics-to-word speech recognition,'' in
  \emph{ICASSP}, 2019.

\bibitem{kahn2020libri}
J.~Kahn, M.~Rivi{\`e}re, W.~Zheng, E.~Kharitonov, Q.~Xu, P.-E. Mazar{\'e},
  J.~Karadayi, V.~Liptchinsky, R.~Collobert, C.~Fuegen \emph{et~al.},
  ``Libri-light: A benchmark for asr with limited or no supervision,'' in
  \emph{ICASSP}, 2020.

\bibitem{lugosch2019speech}
L.~Lugosch, M.~Ravanelli, P.~Ignoto, V.~S. Tomar, and Y.~Bengio, ``Speech model
  pre-training for end-to-end spoken language understanding,'' in
  \emph{InterSpeech}, 2019.

\end{thebibliography}


\end{document}